\documentclass[]{template}

\usepackage[utf8]{inputenc}             
\usepackage[T1]{fontenc}                
\usepackage{url}                        
\usepackage{booktabs}                   
\usepackage{multirow}
\usepackage{colortbl}
\usepackage{multicol}
\usepackage{amsfonts}                   
\usepackage{nicefrac}                   
\usepackage{microtype}                  
\usepackage[dvipsnames]{xcolor}         

\usepackage{latexsym}

\usepackage{graphicx}
\usepackage{float}
\usepackage{subcaption}
\usepackage{wrapfig}
\usepackage{lipsum}

\usepackage{bm}

\usepackage{tabularx} 
\usepackage{ragged2e} 
\newcolumntype{L}{>{\RaggedRight\hangafter=1\hangindent=0em}X}

\usepackage{enumitem}

\usepackage{amsmath}
\usepackage{amssymb}
\usepackage{mathtools}
\usepackage{amsthm}

\setboolean{logo}{true}    

\usepackage[linesnumbered,ruled,vlined]{algorithm2e}

\hypersetup{
    colorlinks=true,
    linkcolor=red,
    citecolor=Cerulean,
    filecolor=magenta,      
    urlcolor=magenta,
}

\usepackage[capitalize,noabbrev]{cleveref}
\crefname{section}{§}{§§}
\Crefname{section}{§}{§§}

\usepackage{calligra}
\DeclareMathAlphabet{\mathcalligra}{T1}{calligra}{m}{n}

\usepackage{pifont}

\theoremstyle{plain}

\theoremstyle{definition}

\theoremstyle{remark}

\renewcommand{\paragraph}[1]{\vspace{1mm}\noindent\textbf{#1}}

\DeclareCaptionLabelFormat{cont}{#1~#2\alph{ContinuedFloat}}
\usepackage[most]{tcolorbox}
\tcbset{
  promptbox/.style={
    top=10pt,
    colback=lightgray!20,
    colframe=Black,
    colbacktitle=NavyBlue,
    enhanced,
    center,
    attach boxed title to top center={yshift=-0.1in,xshift=0.0in},
    boxed title style={boxrule=0pt,colframe=white,},
  }
}
\newtcolorbox{promptbox}[2][]{promptbox, title=#2,#1}
\tcbset{
  takeawaybox/.style={
    top=10pt,
    colback=lightgray!20,
    colframe=Black,
    colbacktitle=BurntOrange,
    enhanced,
    center,
    attach boxed title to top center={yshift=-0.1in,xshift=0.0in},
    boxed title style={boxrule=0pt,colframe=white,},
  }
}
\newtcolorbox{takeawaybox}[2][]{takeawaybox, title=#2,#1}
\tcbset{
  observationbox/.style={
    top=10pt,
    colback=lightgray!20,
    colframe=Black,
    colbacktitle=YellowGreen,
    enhanced,
    center,
    attach boxed title to top center={yshift=-0.1in,xshift=0.0in},
    boxed title style={boxrule=0pt,colframe=white,},
  }
}
\newtcolorbox{observationbox}[2][]{observationbox, title=#2,#1}

\usepackage{xspace}

\newcommand\blfootnote[1]{%
  \begingroup
  \renewcommand\thefootnote{}\footnote{#1}%
  \addtocounter{footnote}{-1}%
  \endgroup
}

\usepackage{CJK}

\usepackage{tabularx}
\usepackage{booktabs}
\usepackage{amsmath}
\usepackage{amssymb}
\usepackage{tcolorbox}
\usepackage{listings}
\usepackage{xcolor}
\usepackage{afterpage}

\usepackage{stfloats}

\title{Timely Machine: Awareness of Time Makes Test-Time Scaling Agentic}

\author[1,2]{Yichuan Ma}
\author[2,3$\dagger$]{Linyang Li}
\author[2]{Yongkang chen}
\author[1,2]{Peiji Li}
\author[2,4]{Xiaozhe Li}
\author[2]{Qipeng Guo}
\author[3]{Dahua Lin}
\author[2]{Kai Chen}

\affil[1]{Fudan University}
\affil[2]{Shanghai AI Laboratory}
\affil[3]{The Chinese University of Hong Kong}
\affil[4]{Tongji University}

\begin{abstract}
As large language models (LLMs) increasingly tackle complex reasoning tasks, test-time scaling has become critical for enhancing capabilities. However, in agentic scenarios with frequent tool calls, the traditional generation-length-based definition breaks down: tool latency decouples inference time from generation length. We propose \textsc{Timely Machine}, redefining test-time as wall-clock time, where models dynamically adjust strategies based on time budgets.
We introduce Timely-Eval, a benchmark spanning high-frequency tool calls, low-frequency tool calls, and time-constrained reasoning. By varying tool latency, we find smaller models excel with fast feedback through more interactions, while larger models dominate high-latency settings via superior interaction quality. Moreover, existing models fail to adapt reasoning to time budgets.
We propose Timely-RL to address this gap. After cold-start supervised fine-tuning, we use reinforcement learning to enhance temporal planning. Timely-RL improves time budget awareness and consistently boosts performance across Timely-Eval. We hope our work offers a new perspective on test-time scaling for the agentic era.
\end{abstract}

\begin{document}

\blfootnote{$\dagger$ Corresponding authors: Linyang Li (lilinyang@pjlab.org.cn)}


\maketitle

\section{Introduction}

As the parameter counts and data volumes of Large Language Models (LLMs) continue to expand, relying solely on scaling these dimensions is becoming increasingly expensive. A new wave of models, represented by OpenAI-o1 and DeepSeek-R1, has pivoted towards exploring test-time scaling, which allocates increased inference time to tackle complex reasoning tasks~\cite{deepseekr1, openaio1}. Drawing inspiration from Group Relative Policy Optimization (GRPO), methods based on Reinforcement Learning with Verifiable Rewards (RLVR) have emerged as a mainstream paradigm for enhancing LLM reasoning~\cite{deepseekmath, qwen2_5_coder, dapo}. 

Furthermore, a significant portion of research applies RLVR to scenarios involving external tools and environmental interactions.~\cite{toolrl, torl}. Empowered by the reasoning capabilities of LLMs, agent systems designed for interactive tasks, such as web search and automated machine learning model training, have also demonstrated substantial application potential~\cite{aide, searchr1}.

Regarding the reasoning capabilities of LLMs, test-time scaling stands as a pivotal concept. Prior to the widespread adoption of LLMs in tool-use scenarios, test-time scaling in standard reasoning tasks (such as mathematics, coding, and commonsense reasoning) was typically correlated with generation time~\cite{learningwhen2plan, budgetthinker, tokenbudgetaware}. Intuitively, in scenarios involving only model inference, an inherent positive correlation exists between time and generation length. 

However, this relationship breaks down when discussing agentic scenarios characterized by complex tool utilization. In such settings, since the latency of tool execution is unpredictable, the total time required to complete a task cannot be estimated solely based on the model's generation length. A concise reasoning trajectory might consume much time if it involves high-latency tools; and conversely, a longer trace could even end faster given rapid environmental feedback. Consequently, the traditional definition of test-time scaling based on generation length proves inapplicable for agentic tasks. \citet{budgetawaretool} provides models with remaining tool call budget through an external plug-in. While this method is effective, it equivalently treats the budget as the number of tool calls, which still overlooks the latency of the tools themselves.

In this work, we introduce the concept of the \textbf{\textsc{Timely Machine}}. By decoupling test time from generation tokens, we restore the concept of "test time" to its original meaning as a physical quantity: time itself. We argue that the awareness of task duration and the ability to plan for task costs should be intrinsic capabilities of LLMs. 

Specifically, when environmental feedback incurs high latency, the model should adjust its interaction frequency and prioritize the quality of each interaction round to achieve better results; conversely, when the environment responds rapidly, the model should engage in faster interactions to gather more sufficient feedback, thereby exploring for higher payoffs within the same duration. 

To validate this capability, we introduce \textbf{Timely-Eval}, a benchmark designed to evaluate how well models adapt their strategies based on temporal feedback. Within Timely-Eval, we curate three distinct interactive scenarios: high-frequency tool, low-frequency tool, and general reasoning tasks under strict time constraints. Across these tasks, we aim to assess whether models are aware of the concept of "time", and whether they can utilize this awareness to dynamically adjust their reasoning strategies accordingly.

We then evaluate model performance under different settings of tool latencies by adjusting the feedback delays. We find that, on tasks with shorter tool latencies, relatively smaller models sometimes achieve better performance than larger models due to their ability to perform more rounds of interactions. However, as interaction time increases, the speed advantage of small models gradually diminishes, while large models, benefiting from their higher interaction quality, often yield greater returns.

To further train models with this accurate time awareness, we first synthesize a large Timely Reasoning dataset, where each sample strictly adheres to the time constraints for task completion. We then use this dataset to perform supervised fine-tuning (SFT) on base models, obtaining initial points with time awareness. We then propose \textbf{Timely-RL} on these initial points for further enhance. As a variant of GRPO, Timely-RL encourages models to maximally utilize the provided time budget. 
After Timely-RL training, the model demonstrates the ability to effectively plan and utilize time budget.

To summarize, our contributions are as follows:

\begin{enumerate}
    \item We introduce Timely Machine, redefining test-time in agentic scenarios as wall-clock time.
    \item We propose Timely-Eval, a benchmark for evaluating reasoning under varying time budgets.
    \item Through evaluation, we find that: (i) model preference shifts with tool latency, smaller models outperform larger ones under low latency through more interactions, while larger models excel when latency is high (ii) existing models lack the ability to adjust reasoning strategies based on time budgets.
    \item We develop Timely-RL, a reinforcement learning approach that teaches models time-aware reasoning. Timely-RL enables models to develop accurate time budget awareness and adapt their reasoning strategies accordingly.
\end{enumerate}


    
    
    
    
\section{Related Work}

\subsection{Test-Time Scaling for LLM Reasoning} 

In research on enhancing LLM reasoning, 
several early works improved reasoning accuracy by voting on multiple responses~\cite{selfconsistency}. Some also explored the possibility of self-correction~\cite{self-reward, self-eval-23, llmcanselfimprove}. \citet{treeofthought} proposed Tree of Thought, which turns reasoning into a tree structure segmented by steps. Later, some studies investigated applying Monte Carlo Tree Search to LLM sampling to tackle more complex reasoning problems~\cite{rstar-math, alphamath}.

\citet{deepseekmath} proposed Group Reward Policy Optimization (GRPO). GRPO fueled a surge of interest in enhancing model reasoning via Reinforcement Learning with Verifiable Rewards \textbf{(RLVR)}. In general reasoning, RLVR-based post-training strategies have brought remarkable performance gains~\cite{qwen2_5_coder, vlm-r1, kimik2, qwen3}.  

Following RLVR’s success on reasoning tasks, subsequent work explored RL practices that involve interacting with tools~\cite{toolrl, retool, r1seracher, internbootcamp, torl}. These efforts show the promise of applying LLMs’ general reasoning ability to more complex, agentic interactive environments.

\subsection{Budget Control for LLM Reasoning} 

Before agentic RL became a trend, researchers investigated lengthy reasoning chains in general tasks~\cite{optimizingttc, stopoverthinking, efficientllm}. In these settings, budget control is equivalent to controlling output length. Consequently, a line of work focused on regulating generation length~\cite{learningwhen2plan, budgetthinker, tokenbudgetaware}, aiming to make LLMs sensitive to inference budgets for dynamic adjustment. Recently, some studies have further incorporated wall-clock latency into general reasoning via additional designs~\cite{timebill, latencytts}. Transitioning to agentic scenarios, AgentTTS~\citep{agenttts} allocates task budgets based on FLOPs, while \citet{budgetawaretool} introduces an external plug-in to modulate tool-use and reasoning budgets. However, these approaches rely on extra designs. We take a more radical stance: time awareness should be an intrinsic capability.
\section{Preliminary Study}

\subsection{Test-Time Scaling in Agentic Scenarios}

Before presenting our approach, we first clarify what test-time means.
In tasks like math and code, test-time was identical to the model's inference time.
Therefore, for a given reasoning task, suppose the model performs $N$ reasoning steps.
The total time cost $t_{\mathrm{all}}$ can be written as:

\begin{equation}
t_{\mathrm{all}} = \sum_{i=1}^{N} t_{\mathrm{gen}}^{(i)},
\end{equation}
where $t_{\mathrm{gen}}^{(i)}$ denotes the generation time at the $i$-th step. Here, total time equals inference time. Therefore, allocating a larger time budget is equivalent to allocating a larger token budget. However, in an LLM-powered agentic task with tool calls, the total time cost $t_{\mathrm{all}}$ becomes:
\begin{equation}
t_{\mathrm{all}} = \sum_{i=1}^{N} t_{\mathrm{gen}}^{(i)} + \sum_{i=1}^{N} t_{\mathrm{tool}}^{(i)},
\end{equation}
where $t_{\mathrm{tool}}^{(i)}$ denotes the time consumed by the tool call at the $i$-th step. Since tool latency is now part of the overall cost, test-time is no longer directly proportional to the number of generated tokens.
For clarity, we discuss the following three cases:

\subsubsection{Tools Take the Lead}

We first consider the case where the tool latency in each round is much larger than generation latency:
\begin{equation}
t_{\mathrm{tool}}^{(i)} \gg t_{\mathrm{gen}}^{(i)}.
\end{equation}
In this case, generation time has little impact on test-time.
Therefore, under a fixed time budget $t_{\mathrm{budget}}$, the number of interaction rounds $N$ is largely determined by tool latency:
\begin{equation}
N \approx \frac{t_{\mathrm{budget}}}{t_{\mathrm{tool}}}.
\end{equation}
Consequently, the per-round interaction quality becomes the key factor that determines task performance.
For such tasks, the model should plan more carefully in each round.

\subsubsection{Models Take the Lead}

In some scenarios, tool latency is negligible, or the model can even solve the task without any tool use:
\begin{equation}
t_{\mathrm{gen}}^{(i)} \gg t_{\mathrm{tool}}^{(i)}.
\end{equation}
In this case, test-time collapses back to the conventional setting, where the total cost is dominated by the number of generated tokens.


\subsubsection{Meeting in the Middle}

In more general settings, the model's generation latency and the tool-call latency are often of a similar scale, i.e.,
\begin{equation}
t_{\mathrm{tool}}^{(i)} \approx t_{\mathrm{gen}}^{(i)}.
\end{equation}

In this regime, test-time compute is jointly determined by tool latency and generation latency.
We define a per-step latency ratio
\begin{equation}
m_i = \frac{t_{\mathrm{tool}}^{(i)}}{t_{\mathrm{gen}}^{(i)}}.
\end{equation}
Then $t_{\mathrm{tool}}(i) = m_i \, t_{\mathrm{gen}}(i)$, the total cost becomes
\begin{equation}
t_{\mathrm{all}} = \sum_{i=1}^{N} (m_i + 1)\, t_{\mathrm{gen}}^{(i)}.
\end{equation}

Since neither $t_{\mathrm{gen}}^{(i)}$ nor $t_{\mathrm{tool}}^{(i)}$ is known beforehand, $m_i$ can vary substantially across steps.
When $m_i > 1$, tool latency dominates, and the model should spend more effort to improve the quality of each tool call;
When $m_i < 1$, generation dominates, and the model may shorten its reasoning to enable more interaction rounds.
Overall, because $m_i$ is unpredictable, token-based budget control is insufficient. Therefore, agent systems must condition their strategies on \emph{time} and adapt dynamically to the observed latency feedback.

\begin{figure*}[t]
    \centering
    \includegraphics[width=0.75\linewidth]{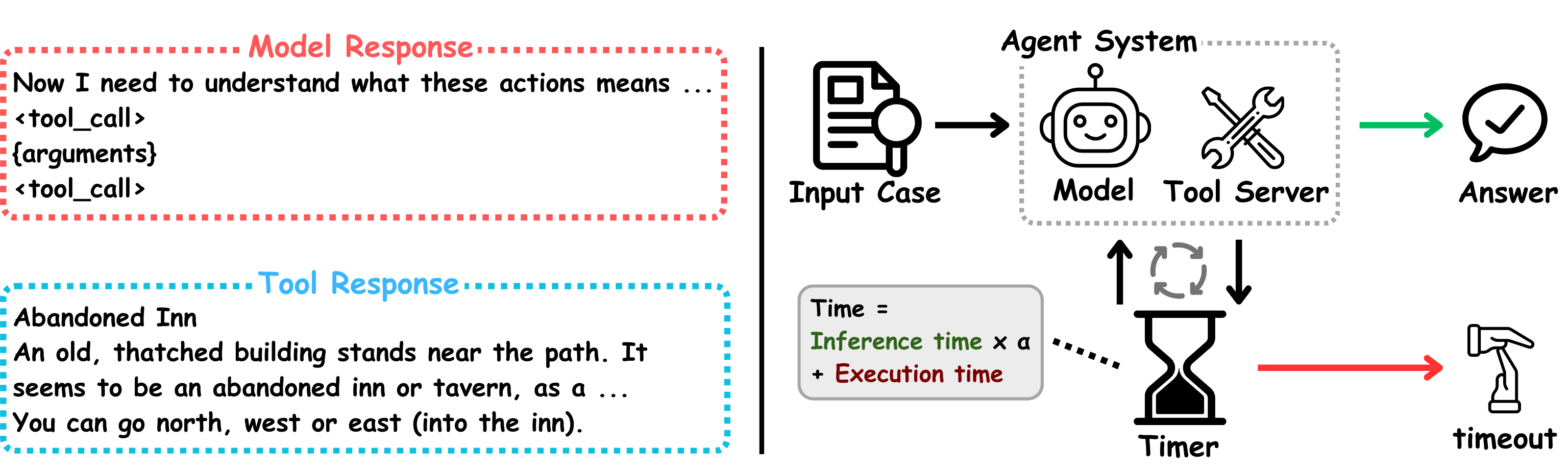}
    \caption{Pipeline for Timely-RL}
    \label{fig:timely-rl}
    \vspace{-2mm}
\end{figure*}

\section{Method}

In this section, we describe \textsc{Timely-Eval} and \textsc{Timely-RL} introduced in the Introduction.

\subsection{Timely-Eval}
To measure an LLM's awareness of time and its ability to regulate behavior in agentic tasks, we introduce \textsc{Timely-Eval}, which covers three task families: (i) Interactive Text Games, (ii) Machine Learning Tasks, and (iii) General Reasoning Tasks.

\subsubsection{Interactive Games}

Interactive fiction games present a natural testbed for agentic scaling.
We build upon \textbf{Jericho}\footnote{\url{https://github.com/microsoft/jericho}}, an interface connecting agents to classic text-based games like \textit{Zork}.
In these environments, models navigate through rounds of interaction, selecting actions to maximize rewards.

Since these games are lightweight, we can easily assign different arbitrary interaction delays to each action without worrying about actual tool execution time. 
This allows us to simulate both high-latency and low-latency scenarios, enabling exploration of how models adapt their strategies across varying time constraints.
We benchmark each model to obtain its average per-step latency $\tau$, then set the time budget as an integer multiple of $\tau$. 


\subsubsection{Machine Learning}

In machine learning tasks, tool execution dominates the time budget.
We adapt several tasks from MLEBench-Lite~\cite{mlebench}, where LLMs should train ML models to solve given problems.
Models should interact with code interpreter, submitting solutions and observing execution outputs, error messages, and performance scores across multiple iterations.

After estimating the average per-round latency $\tau$ (including both LLM reasoning and model training), we allocate multiples of $\tau$ as the time budget.
Because the training of ML models is typically time-consuming, this suite evaluates whether the LLM can make effective use of additional time budget when tool cost dominates.

\subsubsection{General Reasoning}

Finally, we return to standard reasoning tasks, where test-time is equivalent to generation time.
To ensure the model conditions on \emph{time} (rather than only token length), we provide a timer tool that reports elapsed time. Models must query it during reasoning to avoid timing out.
As before, we run speed test to obtain $\tau$ and set per-instance budgets as scaled multiples. This suite tests whether time awareness emerges naturally.

\subsection{Timely Cold Start}

Before applying reinforcement learning, we first distill supervised data to provide a cold start and establish baselines.

On general reasoning tasks, We distill 1M training examples from Qwen3-235B-2507-Instruct~\cite{qwen3} on queries sampled from AM-DeepSeek-R1-Distilled-1.4M~\cite{AM1.4M}. Specifically, we equip the teacher model with a timer tool, enabling it to produce solutions under varying time budgets.

Similarly, we distill interaction traces from DeepSeek-V3.2~\cite{deepseekv3.2} on interactive games.
To prevent memorization, we limit distillation to the first \textbf{50} steps per game, ensuring that subsequent exploration must be learned rather than merely recalled.
We do not perform task-specific distillation for ML tasks.

After distillation, we combine the synthetic dataset with DataScience-Instruct-500K~\cite{deepanalyze} to form the final cold-start mixture.
Through SFT on various base models, we provide the initialization point for subsequent RL training.

\subsection{Timely-RL}
\label{sec:timelyrl}

Tool latency creates dynamic, real-time feedback signals that naturally lend themselves to reinforcement learning. We therefore propose Timely-RL, a time-aware RL approach that trains models to optimize their behavior under time constraints. The whole pipeline is shown in Fig.~\ref{fig:timely-rl}.

In each interaction round $i$, the time feedback $t^{(i)}$ 
is computed as the sum of inference time $t_{\text{gen}}^{(i)}$ and tool execution time $t_{\text{tool}}^{(i)}$. 
To adapt the model to different conditions, we multiply $t_{\text{gen}}^{(i)}$ by a coefficient $\alpha$ as the feedback received by the model, simulating scenarios on faster or slower devices. 

Next, we design reward functions for the three tasks during RL training. For each response, let $r$ denote the accuracy reward, $t$ denote the time cost for task completion, and $T_{\text{max}}$ denote the time limit. The reward function can be expressed as:
\begin{equation}
R(t, r) = 
\begin{cases}
0, & t > T_{\max} \\
r_{\text{f}}, & t \le T_{\max}, r = 0 \\
r_{\text{f}} + r + \lambda U(t), & t \le T_{\max}, r > 0
\end{cases}
\end{equation}
where $r_{\text{f}}$ is a fixed reward for correct formatting and timely completion, and $U(t)$ represents the model's utilization rate of time cost, calculated as:
\begin{equation}
U(t) = \sin\left( \frac{\pi}{2} \cdot \min\left(\frac{t}{T_{\text{max}}}, 1\right) \right)
\end{equation}

This ensures that as the time ratio nears 1.0, the reward increases more slowly, preventing minor time differences near the deadline from causing large reward fluctuations.

\begin{figure*}[t]
    \centering
    \includegraphics[width=1\linewidth]{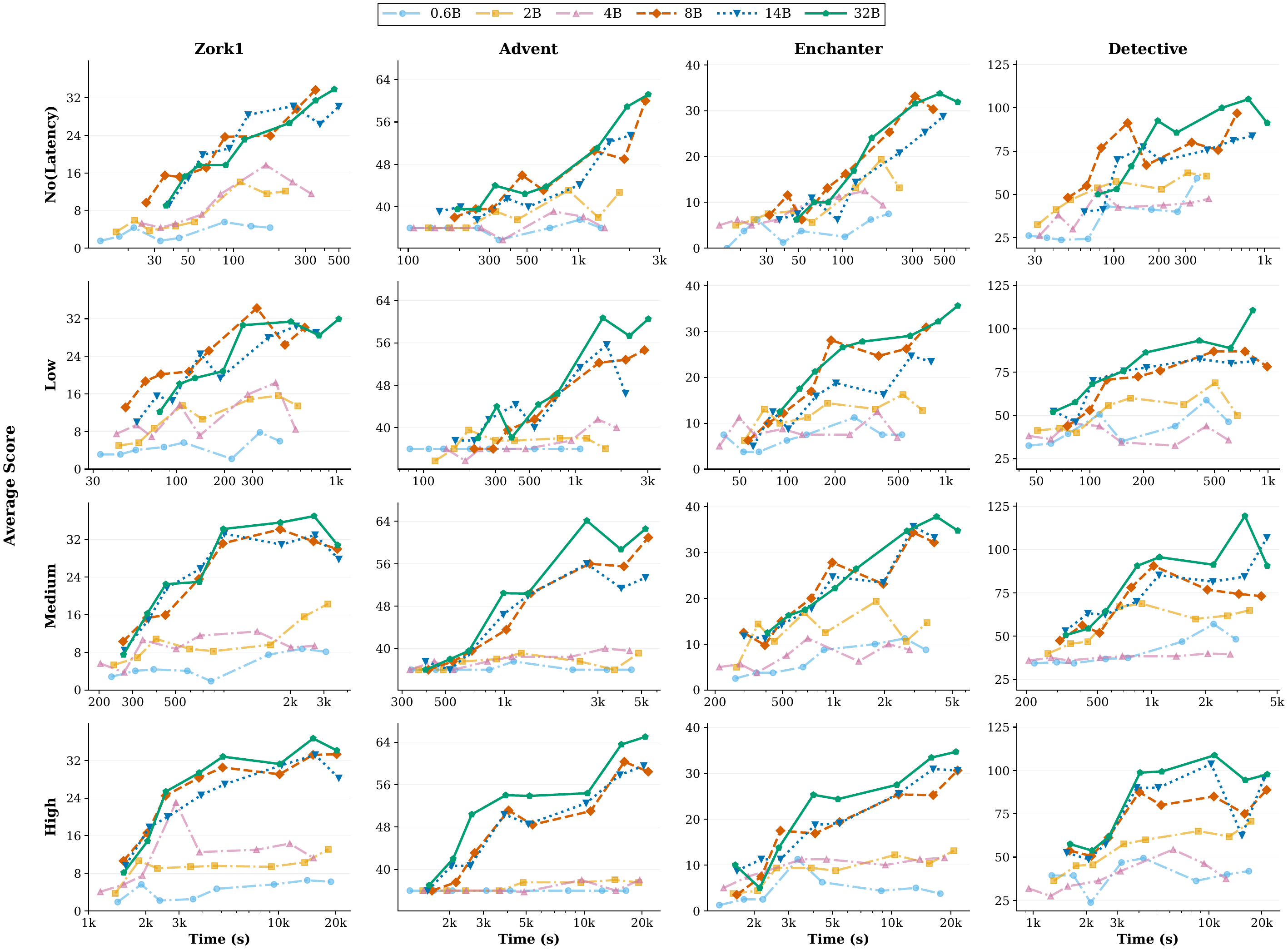}
    \caption{Scaling experiments across four text-based interactive games (Zork1, Advent, 
Enchanter, Detective) using Qwen3 models (0.6B-32B parameters). All models are 
cold-started with the same dataset. Each row represents different tool call 
latency settings: No latency, Low ($\sim$2s), Medium ($\sim$10s), and High ($\sim$50s).}
    \label{fig:time_scaling_exp}
    \vspace{-2mm}
\end{figure*}
\section{Experiments}


\subsection{Tool Latency Flips the Scaling Winner}

\label{sec:scaling_exp}

\subsubsection{8B Beats 32B}

To compare the scaling performance of models with different sizes under various tool call scenarios, we select four well-known interactive text games as evaluation benchmarks: Enchanter, Zork1, Advent, and Detective. We perform cold start on six models from the Qwen3 series, ranging from 0.6B to 32B parameters, and conduct evaluations under four settings with different tool call latencies, as shown in Fig.~\ref{fig:time_scaling_exp}. 
The deployment resources are fixed for all models. 

The results show that when tool latency is low, smaller models can outperform larger ones within the same time constraint due to faster inference. For instance,
in the zero latency setting on \textbf{Detective}, with a 50s-120s time budget, the 8B model achieves top performance among all models, outperforming both 14B and 32B. On other games, the largest 32B model still shows no significant performance advantage under the same budget.


\begin{table*}[t]
\centering
\caption{Evaluation results on Timely Eval across three task categories. For general reasoning tasks, we report the average accuracy under various time budgets. For interactive games, we show the average game scores achieved under different budgets (8 trials per game). For ML tasks, we present the average test set accuracy of generated code (8 trials per task). The best score is in \textbf{bold}, while the second best is \underline{underlined}.} 
\label{tab:Timely_Eval_main_table}
\footnotesize  
\setlength{\tabcolsep}{5.5pt}  
\renewcommand{\arraystretch}{1.1}
\begin{tabular}{l|ccc|cccc|cccc}
\toprule
\multirow{3}{*}{\textbf{Model}} & \multicolumn{3}{c|}{\textbf{General Reasoning}} & \multicolumn{4}{c|}{\textbf{Interactive Reasoning}} & \multicolumn{4}{c}{\textbf{Machine Learning}} \\
\cmidrule(lr){2-4} \cmidrule(lr){5-8} \cmidrule(lr){9-12}
& MATH & AIME & GPQA & Zork1 & Advent & Enchan & Detect & Leaf & Space & Pizza & Insult \\
& & & Diam. & & & -ter & -ive & Clf. & -ship & & Detect \\
\midrule
GPT-5.1(medium) & 71.5 & \textbf{46.7} & \textbf{71.0} & \underline{34.1} & \textbf{57.6} & 24.4 & \textbf{105.0} & 67.5 & 63.5 & 58.0 & 79.0 \\
DeepSeek-V3.2 & \textbf{82.9} & 43.3 & 58.7 & 24.9 & 48.7 & 15.9 & 63.2 & 72.0 & 68.1 & 68.6 & 78.0 \\
Gemini2.5-pro & 63.0 & 37.5 & \underline{59.0} & \textbf{34.9} & \underline{50.7} & \textbf{34.1} & 71.9 & 71.4 & 65.4 & 74.6 & 75.1 \\
\midrule
Qwen3-8B & 71.2 & 40.0 & 37.5 & 2.3 & 36.0 & 5.2 & 54.1 & 32.0 & 17.8 & 28.6 & 51.9 \\
Qwen3-14B & 70.8 & 41.7 & 21.0 & 9.8 & 34.9 & 9.5 & 50.5 & 46.4 & 55.9 & 60.7 & 70.6 \\
Qwen3-32B & 75.0 & \underline{45.7} & 35.5 & 14.4 & 38.2 & 11.7 & 70.0 & 51.8 & 50.1 & 59.0 & 57.2 \\
\midrule
TimelyLM-8B & \underline{78.0} & 42.5 & 49.5 & 27.5 & 48.5 & \underline{29.5} & \underline{88.1} & \textbf{93.9} & \textbf{74.4} & \textbf{76.1} & \textbf{84.0} \\
\bottomrule
\end{tabular}
\end{table*}

However, as tool latency increases, the task transitions to a tool-dominated regime. Under the same time budget, different-sized models perform a similar number of interaction rounds. At this point, larger models demonstrate higher interaction quality.
In the high latency setting, 32B consistently achieves the best performance across all tasks. Additionally, when model size is too small, exemplified by the 0.6B, 1.7B, and 4B models, despite their fast generation speed, their performance remains inferior to larger models regardless of tool setting due to capability limitations. Therefore, for an agentic task, beyond tool latency, the model's inherent capability must also be considered. An appropriate model selection should balance inference speed and performance.

\subsection{Experimental Setups}

\subsubsection{Timely-Eval Details}

We first introduce the components of Timely-Eval. For interactive games, We collect and wrap 57 text-based interactive games from the Jericho environment. Models are provided with tools to execute actions, query valid moves, check scores, and terminate games.

General benchmarks comprise three reasoning benchmarks: AIME, MATH, and GPQA-diamond ~\cite{math, gpqa}. Models are provided with a tool to track elapsed inference time, and must adjust their reasoning strategies based on time costs and budget consumption.

ML benchmarks contain four ML tasks: Leaf Classification, Spaceship Titanic, Random Acts of Pizza, and Detecting Insults in Social Commentary. Models are allowed to use tools to execute code, obtain results, and monitor elapsed time.

\subsubsection{Evaluation Setups}

We then introduce the experimental settings in Timely-Eval. For General Benchmarks, we set time budgets at 0.75$\times$, 1.0$\times$, 2.0$\times$, and 3.0$\times$ of this baseline, treating responses exceeding the budget as incorrect. We report accuracy averaged across budgets.

For Machine Learning Benchmarks, we measure the average time per reasoning step (including tool calls) and allocate budgets ranging from 1$\times$ to 5$\times$ this duration.

For Interactive Games, we assign latencies to different tools to simulate real execution. We evaluate on four games (Enchanter, Zork1, Advent, Detective) with budgets equivalent to 30, 50, 100, and 200 steps, reporting scores averaged across settings. In Section~\ref{sec:scaling_exp}, budgets range from 30 to over 400 steps.

\subsubsection{Training Setups}

We first perform SFT as cold-start, followed by Timely-RL. SFT runs for 2 epochs using Xtuner~\cite{xtuner}. For RL, each sample is rolled out 8 times with batch size 16,
with maximum generation length of 16,384 tokens. We set learning rate to 1e-6 and clip ratio to [0.2, 0.28]. We use VeRL for RL~\cite{verl}.

\subsection{Timely-Eval Results}

\begin{table*}[t]
\centering
\footnotesize  
\setlength{\tabcolsep}{4.4pt}  
\renewcommand{\arraystretch}{1.15}
\caption{Performance under a 0.75$\times$ time budget. \textbf{On Time}: percentage of samples finished within the time limit (\%). \textbf{ACC}: accuracy (\%). \textbf{Tool-use rate}: percentage of samples that invoked at least one tool call (\%).}
\label{tab:ontime_075}
\begin{tabular}{lccc|ccc|ccc}
\toprule
& \multicolumn{3}{c|}{\textbf{AIME} (0.75$\times$)} 
& \multicolumn{3}{c|}{\textbf{GPQA-Diamond} (0.75$\times$)} 
& \multicolumn{3}{c}{\textbf{MATH} (0.75$\times$)} \\
\textbf{Model} 
& \textbf{On Time} & \textbf{ACC} & \textbf{Tool-use rate}
& \textbf{On Time} & \textbf{ACC} & \textbf{Tool-use rate}
& \textbf{On Time} & \textbf{ACC} & \textbf{Tool-use rate} \\
\midrule
DeepSeek-V3.2     & \underline{33.3} & \underline{33.3} & \underline{100.0} & \underline{43.4} & 33.3 & \underline{100.0} & \textbf{75.0} & \textbf{74.0} & \underline{100.0} \\
GPT5.1            & 33.3 & 30.0 & 40.0  & 48.0 & \textbf{44.0} & 8.0   & 34.0 & 34.0 & 8.0   \\
\midrule
Qwen3-8B          & 30.0 & 26.7 & 53.3  & 36.0 & 20.0 & 94.0  & 48.0 & 42.0 & 66.0  \\
Qwen3-14B         & 30.0 & 20.0 & 10.0   & 18.0 & 12.0 & 8.0   & 38.0 & 32.0 & 6.0   \\
Qwen-32B          & 33.3 & 26.7 & 26.7  & 34.0 & 28.0 & 52.0  & 46.0 & 42.0 & 26.0  \\
\midrule
TimelyLM-8B & \textbf{53.3} & \textbf{33.3} & \textbf{100.0} & \textbf{52.0} & \underline{36.0} & \textbf{100.0} & \underline{56.0} & \underline{52.0} & \textbf{100.0} \\
\bottomrule
\end{tabular}
\end{table*}

\begin{figure*}[t]
    \centering
    \vspace{0.5em}
    \includegraphics[width=0.95\linewidth]{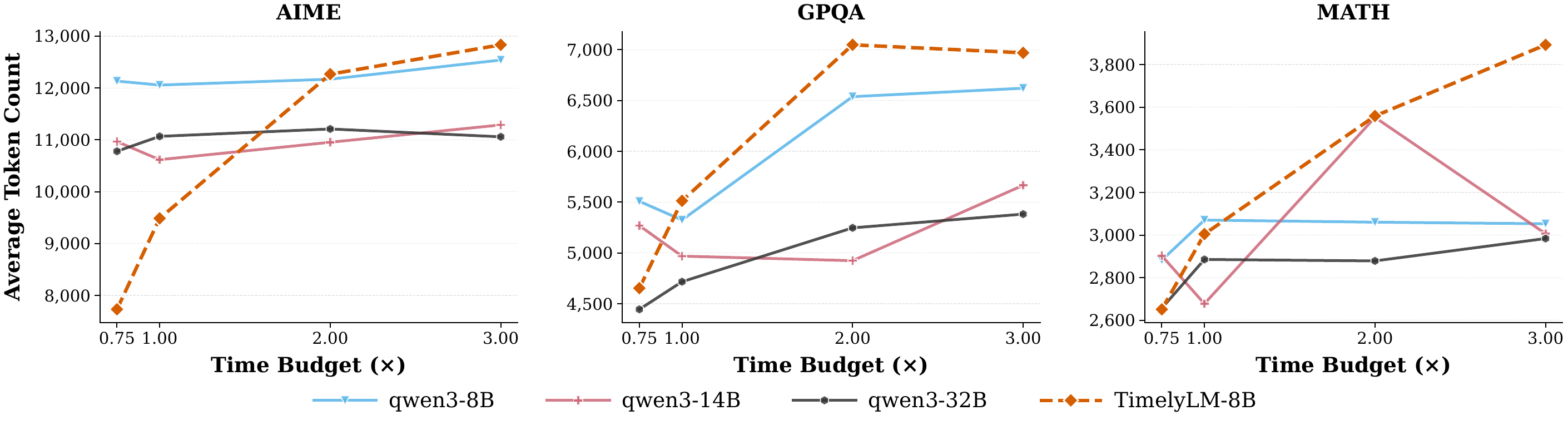}
    \vspace{-0.5em}
    \caption{Changes in reasoning length across different time budgets. The time budget is the model's original reasoning time scaled by factors such as $0.75\times$ and $1.0\times$.}
    \label{fig:token_curve}
    \vspace{-0.5em}
\end{figure*}

Starting from the cold-started 8B, we apply Timely-RL across three tasks to obtain TimelyLM-8B. For evaluation on Timely-Eval, we select several mainstream general models as performance references, and also evaluate models from the Qwen3 series for direct comparison. The evaluation results on Timely-Eval are shown in Tab.~\ref{tab:Timely_Eval_main_table}.

First, TimelyLM-8B outperforms Qwen3-8B on all three general benchmarks. 
Notably, all models show significant performance drops on Timely-Eval compared to raw benchmarks. 
For reasoning models(such as DeepSeek-V3.2), this is because they cannot control the generation length, and thus frequently fail to complete the task within the time budget at 0.75$\times$ or even larger time constraints. While GPT-5.1 does not perform lengthy reasoning, the time constraint forces it into extremely short reasoning, leading to performance degradation. The Qwen series models consider fewer tool calls during reasoning. Due to their faster inference speed, they occasionally complete reasoning within the time limit. More detailed discussion of this phenomenon is provided in Section~\ref{sec:on_time}.

On Interactive Games and Machine Learning tasks, TimelyLM-8B achieves strong performance. On ML tasks, larger models sometimes choose to implement more complex solutions, causing the code execution to exceed the time limit. TimelyLM, however, produces more stable code implementations. On Interactive Games, although we only use the first 50 steps as a cold start, TimelyLM still achieves performance comparable to the strongest general models, demonstrating that the model has effectively learned exploration strategies for these interactive games.

\subsection{Training LLMs to be ON TIME}
\label{sec:on_time}

Furthermore, we design additional experiments to verify whether Timely-RL truly  equips the model with better time awareness. 

\begin{figure*}[t]
    \centering
    \includegraphics[width=1.0\linewidth]{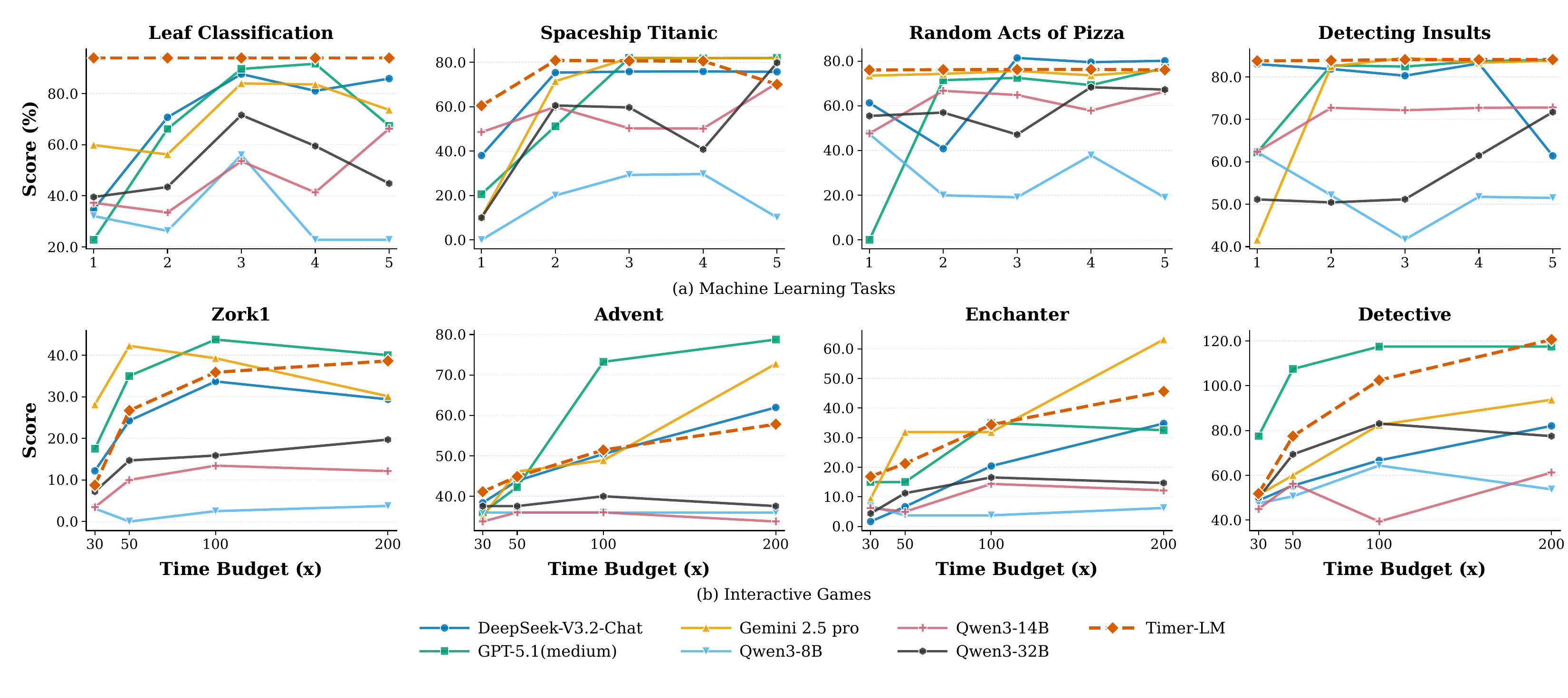}
    \vspace{-2em}
    \caption{Scaling Experiments on ML tasks and Interactive Games.}
    \label{fig:combined_scaling}
    \vspace{-1em}
\end{figure*}


\subsubsection{Behavior under reduced budgets}

To demonstrate this, we first select general benchmarks and compare the on-time completion rates of different models under a 0.75$\times$ time budget. As shown in Tab.~\ref{tab:ontime_075}, models exemplified by Qwen3-8B cannot adapt to shorter time constraints and thus fail to complete reasoning within limit. However, TimelyLM-8B can leverage the timing tool to obtain the remaining budget in real-time and adjust its reasoning process across the tasks, thereby enabling more tasks to be completed on time.

Additionally, we observe two interesting phenomena. First, the performance variation caused by budget reduction differs across benchmarks. On more challenging tasks like AIME and GPQA-diamond, higher completion rates do not lead TimelyLM to significantly outperform other models. However, on simpler problems like MATH, the model's on-time completion rate more closely aligns with its accuracy rate, meaning that most tasks completed on time are solved correctly. On such tasks, the benefit of \textbf{completing tasks on time} outweighs that of performing more extensive reasoning.

Second, comparing more powerful models, DeepSeek-V3.2 better utilizes tool calls to obtain remaining time, while GPT-5.1 prefers single-round reasoning to meet time constraints. A similar pattern appears between Qwen3-8B and Qwen3-32B, 32B prefers single-round reasoning followed by direct summarization, while 8B is more willing to invoke tool calls before summarizing. The 14B model almost completely ignores the instruction requirements for tool calls, instead performing complete single-round reasoning directly.



\subsubsection{Behavior under increased budgets}
Moreover, Fig~\ref{fig:token_curve} shows the changes in reasoning length under different time budgets. We observe that TimelyLM exhibits a significant increasing trend in reasoning length as the time budget increases. In contrast, the Qwen series models do not show this trend, and their reasoning length increases marginally under different time budgets. This demonstrates that Timely-RL indeed enables the model to learn to adjust reasoning strategies based on changing time budgets.

\subsection{Task-dependent budget sensitivity}

Fig.~\ref{fig:combined_scaling} illustrates the impact of providing additional time budgets on performance for interactive games and ML tasks. On ML tasks, we observe little direct relationship between increased budgets and performance improvements. The first few rounds of generated code often determine the final performance. In contrast, on Interactive Games, additional time budgets lead to clear score improvements. 

Therefore, tasks exemplified by Machine Learning rely more heavily on generation quality. Once a solution is good enough, further test-time scaling yields diminishing returns. Conversely, Interactive Games require multiple rounds of interaction to obtain rewards. For such exploratory tasks, the benefits of continued test-time scaling remain evident.

\subsection{Ablation Study on Timely-RL}

\begin{table}[t]
\centering
\small
\begin{tabular}{lccc}
\toprule
\textbf{Method} & \textbf{AIME} & \textbf{GPQA} & \textbf{MATH} \\
\midrule
Qwen3-8B & 30.0 & 36.0 & 48.0 \\
\quad + SFT & 46.7 & 48.0 & 56.0 \\
\quad + GRPO & 33.3 & 46.0 & 48.0 \\
\quad + Timely-RL & \textbf{53.3} & \textbf{52.0} & \textbf{56.0} \\
\bottomrule
\end{tabular}
\caption{On-time completion rates under 0.75$\times$ time budget. All methods start from Qwen3-8B.}
\label{tab:ablation}
\end{table}

Finally, Table~\ref{tab:ablation} presents our ablation study on different stages of Timely-RL training. The results demonstrate that Timely-RL achieves notably higher on-time completion rates compared to SFT alone. Moreover, when applying standard RLVR training without time constraints on the SFT model, the model fails to learn comparable reasoning strategies. Therefore, we conclude that Timely-RL training effectively enhances the model's awareness of task time budgets.

\section{Conclusion}


In this work, we propose the concept of Timely Machine, arguing that test time in agentic scenarios should directly correspond to physical time itself. We advocate that LLMs should possess awareness of time budgets. To this end, we introduce Timely-RL to train models to better perceive task execution time and dynamically adjust their reasoning accordingly. We hope our work provides a new perspective for test-time scaling in agentic scenarios.

\section*{Limitations}

Finally, we discuss several limitations of this work. First, our evaluation is currently conducted on text-based tasks. In multimodal interactive tasks, processing image or video streams typically introduces more complex and variable latencies, presenting a more challenging scenario. Second, our framework does not fully address cases where LLMs are deployed in multi-agent systems. In these multi-agent settings, latency may arise from another LLM's generation time, and training LLMs for effective collaboration in such contexts remains an open topic. Finally, an interesting observation is that while models become more timely under time constraints, accuracy improvements vary with benchmark difficulty. Future work could explore how to better balance the trade-off between time constraints and model performance.

\clearpage
\bibliographystyle{plainnat}
\bibliography{refs}

@article{selfconsistency,
  title={Self-consistency improves chain of thought reasoning in language models},
  author={Wang, Xuezhi and Wei, Jason and Schuurmans, Dale and Le, Quoc and Chi, Ed and Narang, Sharan and Chowdhery, Aakanksha and Zhou, Denny},
  journal={arXiv preprint arXiv:2203.11171},
  year={2022}
}

@article{treeofthought,
  title={Tree of thoughts: Deliberate problem solving with large language models},
  author={Yao, Shunyu and Yu, Dian and Zhao, Jeffrey and Shafran, Izhak and Griffiths, Tom and Cao, Yuan and Narasimhan, Karthik},
  journal={Advances in neural information processing systems},
  volume={36},
  pages={11809--11822},
  year={2023}
}

@article{rstar-math,
  title={rStar-Math: Small LLMs Can Master Math Reasoning with Self-Evolved Deep Thinking},
  author={Guan, Xinyu and Zhang, Li Lyna and Liu, Yifei and Shang, Ning and Sun, Youran and Zhu, Yi and Yang, Fan and Yang, Mao},
  journal={arXiv preprint arXiv:2501.04519},
  year={2025}
}

@article{alphamath,
  title={Alphamath almost zero: process supervision without process},
  author={Chen, Guoxin and Liao, Minpeng and Li, Chengxi and Fan, Kai},
  journal={Advances in Neural Information Processing Systems},
  volume={37},
  pages={27689--27724},
  year={2024}
}

@article{deepseekmath,
  title={Deepseekmath: Pushing the limits of mathematical reasoning in open language models},
  author={Shao, Zhihong and Wang, Peiyi and Zhu, Qihao and Xu, Runxin and Song, Junxiao and Bi, Xiao and Zhang, Haowei and Zhang, Mingchuan and Li, YK and Wu, Yang and others},
  journal={arXiv preprint arXiv:2402.03300},
  year={2024}
}

@inproceedings{self-reward,
  title={Self-rewarding language models},
  author={Yuan, Weizhe and Pang, Richard Yuanzhe and Cho, Kyunghyun and Li, Xian and Sukhbaatar, Sainbayar and Xu, Jing and Weston, Jason E},
  booktitle={Forty-first International Conference on Machine Learning},
  year={2024}
}

@inproceedings{self-eval-23,
  title={Self-evaluation improves selective generation in large language models},
  author={Ren, Jie and Zhao, Yao and Vu, Tu and Liu, Peter J and Lakshminarayanan, Balaji},
  booktitle={Proceedings on},
  pages={49--64},
  year={2023},
  organization={PMLR}
}

@inproceedings{llmcanselfimprove,
  title={Large language models can self-improve},
  author={Huang, Jiaxin and Gu, Shixiang and Hou, Le and Wu, Yuexin and Wang, Xuezhi and Yu, Hongkun and Han, Jiawei},
  booktitle={Proceedings of the 2023 conference on empirical methods in natural language processing},
  pages={1051--1068},
  year={2023}
}

@article{qwen2_5_coder,
  title={Qwen2. 5-coder technical report},
  author={Hui, Binyuan and Yang, Jian and Cui, Zeyu and Yang, Jiaxi and Liu, Dayiheng and Zhang, Lei and Liu, Tianyu and Zhang, Jiajun and Yu, Bowen and Lu, Keming and others},
  journal={arXiv preprint arXiv:2409.12186},
  year={2024}
}

@article{vlm-r1,
  title={Vlm-r1: A stable and generalizable r1-style large vision-language model},
  author={Shen, Haozhan and Liu, Peng and Li, Jingcheng and Fang, Chunxin and Ma, Yibo and Liao, Jiajia and Shen, Qiaoli and Zhang, Zilun and Zhao, Kangjia and Zhang, Qianqian and others},
  journal={arXiv preprint arXiv:2504.07615},
  year={2025}
}

@article{kimik2,
  title={Kimi k2: Open agentic intelligence},
  author={Team, Kimi and Bai, Yifan and Bao, Yiping and Chen, Guanduo and Chen, Jiahao and Chen, Ningxin and Chen, Ruijue and Chen, Yanru and Chen, Yuankun and Chen, Yutian and others},
  journal={arXiv preprint arXiv:2507.20534},
  year={2025}
}

@article{qwen3,
  title={Qwen3 technical report},
  author={Yang, An and Li, Anfeng and Yang, Baosong and Zhang, Beichen and Hui, Binyuan and Zheng, Bo and Yu, Bowen and Gao, Chang and Huang, Chengen and Lv, Chenxu and others},
  journal={arXiv preprint arXiv:2505.09388},
  year={2025}
}

@article{toolrl,
  title={Toolrl: Reward is all tool learning needs},
  author={Qian, Cheng and Acikgoz, Emre Can and He, Qi and Wang, Hongru and Chen, Xiusi and Hakkani-T{\"u}r, Dilek and Tur, Gokhan and Ji, Heng},
  journal={arXiv preprint arXiv:2504.13958},
  year={2025}
}

@article{retool,
  title={Retool: Reinforcement learning for strategic tool use in llms},
  author={Feng, Jiazhan and Huang, Shijue and Qu, Xingwei and Zhang, Ge and Qin, Yujia and Zhong, Baoquan and Jiang, Chengquan and Chi, Jinxin and Zhong, Wanjun},
  journal={arXiv preprint arXiv:2504.11536},
  year={2025}
}

@article{r1seracher,
  title={R1-searcher: Incentivizing the search capability in llms via reinforcement learning},
  author={Song, Huatong and Jiang, Jinhao and Min, Yingqian and Chen, Jie and Chen, Zhipeng and Zhao, Wayne Xin and Fang, Lei and Wen, Ji-Rong},
  journal={arXiv preprint arXiv:2503.05592},
  year={2025}
}

@article{internbootcamp,
  title={Internbootcamp technical report: Boosting llm reasoning with verifiable task scaling},
  author={Li, Peiji and Ye, Jiasheng and Chen, Yongkang and Ma, Yichuan and Yu, Zijie and Chen, Kedi and Cui, Ganqu and Li, Haozhan and Chen, Jiacheng and Lyu, Chengqi and others},
  journal={arXiv preprint arXiv:2508.08636},
  year={2025}
}

@article{torl,
  title={Torl: Scaling tool-integrated rl},
  author={Li, Xuefeng and Zou, Haoyang and Liu, Pengfei},
  journal={arXiv preprint arXiv:2503.23383},
  year={2025}
}

@article{optimizingttc,
  title={Optimizing test-time compute via meta reinforcement fine-tuning},
  author={Qu, Yuxiao and Yang, Matthew YR and Setlur, Amrith and Tunstall, Lewis and Beeching, Edward Emanuel and Salakhutdinov, Ruslan and Kumar, Aviral},
  journal={arXiv preprint arXiv:2503.07572},
  year={2025}
}

@article{stopoverthinking,
  title={Stop overthinking: A survey on efficient reasoning for large language models},
  author={Sui, Yang and Chuang, Yu-Neng and Wang, Guanchu and Zhang, Jiamu and Zhang, Tianyi and Yuan, Jiayi and Liu, Hongyi and Wen, Andrew and Zhong, Shaochen and Zou, Na and others},
  journal={arXiv preprint arXiv:2503.16419},
  year={2025}
}

@article{efficientllm,
  title={Efficient contextual llm cascades through budget-constrained policy learning},
  author={Zhang, Xuechen and Huang, Zijian and Taga, Ege Onur and Joe-Wong, Carlee and Oymak, Samet and Chen, Jiasi},
  journal={Advances in Neural Information Processing Systems},
  volume={37},
  pages={91691--91722},
  year={2024}
}

@article{learningwhen2plan,
  title={Learning When to Plan: Efficiently Allocating Test-Time Compute for LLM Agents},
  author={Paglieri, Davide and Cupia{\l}, Bart{\l}omiej and Cook, Jonathan and Piterbarg, Ulyana and Tuyls, Jens and Grefenstette, Edward and Foerster, Jakob Nicolaus and Parker-Holder, Jack and Rockt{\"a}schel, Tim},
  journal={arXiv preprint arXiv:2509.03581},
  year={2025}
}

@article{budgetthinker,
  title={Budgetthinker: Empowering budget-aware llm reasoning with control tokens},
  author={Wen, Hao and Wu, Xinrui and Sun, Yi and Zhang, Feifei and Chen, Liye and Wang, Jie and Liu, Yunxin and Liu, Yunhao and Zhang, Ya-Qin and Li, Yuanchun},
  journal={arXiv preprint arXiv:2508.17196},
  year={2025}
}

@inproceedings{tokenbudgetaware,
  title={Token-budget-aware llm reasoning},
  author={Han, Tingxu and Wang, Zhenting and Fang, Chunrong and Zhao, Shiyu and Ma, Shiqing and Chen, Zhenyu},
  booktitle={Findings of the Association for Computational Linguistics: ACL 2025},
  pages={24842--24855},
  year={2025}
}

@article{budgetawaretool,
  title={Budget-Aware Tool-Use Enables Effective Agent Scaling},
  author={Liu, Tengxiao and Wang, Zifeng and Miao, Jin and Hsu, I and Yan, Jun and Chen, Jiefeng and Han, Rujun and Xu, Fangyuan and Chen, Yanfei and Jiang, Ke and others},
  journal={arXiv preprint arXiv:2511.17006},
  year={2025}
}

@article{mlebench,
  title={Mle-bench: Evaluating machine learning agents on machine learning engineering},
  author={Chan, Jun Shern and Chowdhury, Neil and Jaffe, Oliver and Aung, James and Sherburn, Dane and Mays, Evan and Starace, Giulio and Liu, Kevin and Maksin, Leon and Patwardhan, Tejal and others},
  journal={arXiv preprint arXiv:2410.07095},
  year={2024}
}

@misc{AM1.4M,
      title={1.4 Million Open-Source Distilled Reasoning Dataset to Empower Large Language Model Training}, 
      author={Han Zhao and Haotian Wang and Yiping Peng and Sitong Zhao and Xiaoyu Tian and Shuaiting Chen and Yunjie Ji and Xiangang Li},
      year={2025},
      eprint={2503.19633},
      archivePrefix={arXiv},
      primaryClass={cs.CL},
      url={https://arxiv.org/abs/2503.19633}, 
}

@article{deepseekv3.2,
  title={Deepseek-v3. 2: Pushing the frontier of open large language models},
  author={Liu, Aixin and Mei, Aoxue and Lin, Bangcai and Xue, Bing and Wang, Bingxuan and Xu, Bingzheng and Wu, Bochao and Zhang, Bowei and Lin, Chaofan and Dong, Chen and others},
  journal={arXiv preprint arXiv:2512.02556},
  year={2025}
}

@article{deepanalyze,
  title={Deepanalyze: Agentic large language models for autonomous data science},
  author={Zhang, Shaolei and Fan, Ju and Fan, Meihao and Li, Guoliang and Du, Xiaoyong},
  journal={arXiv preprint arXiv:2510.16872},
  year={2025}
}

@article{deepseekr1,
  title={Deepseek-r1: Incentivizing reasoning capability in llms via reinforcement learning},
  author={Guo, Daya and Yang, Dejian and Zhang, Haowei and Song, Junxiao and Zhang, Ruoyu and Xu, Runxin and Zhu, Qihao and Ma, Shirong and Wang, Peiyi and Bi, Xiao and others},
  journal={arXiv preprint arXiv:2501.12948},
  year={2025}
}

@article{openaio1,
  title={Openai o1 system card},
  author={Jaech, Aaron and Kalai, Adam and Lerer, Adam and Richardson, Adam and El-Kishky, Ahmed and Low, Aiden and Helyar, Alec and Madry, Aleksander and Beutel, Alex and Carney, Alex and others},
  journal={arXiv preprint arXiv:2412.16720},
  year={2024}
}

@article{dapo,
  title={Dapo: An open-source llm reinforcement learning system at scale},
  author={Yu, Qiying and Zhang, Zheng and Zhu, Ruofei and Yuan, Yufeng and Zuo, Xiaochen and Yue, Yu and Dai, Weinan and Fan, Tiantian and Liu, Gaohong and Liu, Lingjun and others},
  journal={arXiv preprint arXiv:2503.14476},
  year={2025}
}

@article{searchr1,
  title={Search-r1: Training llms to reason and leverage search engines with reinforcement learning},
  author={Jin, Bowen and Zeng, Hansi and Yue, Zhenrui and Yoon, Jinsung and Arik, Sercan and Wang, Dong and Zamani, Hamed and Han, Jiawei},
  journal={arXiv preprint arXiv:2503.09516},
  year={2025}
}

@article{aide,
  title={Aide: Ai-driven exploration in the space of code},
  author={Jiang, Zhengyao and Schmidt, Dominik and Srikanth, Dhruv and Xu, Dixing and Kaplan, Ian and Jacenko, Deniss and Wu, Yuxiang},
  journal={arXiv preprint arXiv:2502.13138},
  year={2025}
}

@inproceedings{gpqa,
  title={Gpqa: A graduate-level google-proof q\&a benchmark},
  author={Rein, David and Hou, Betty Li and Stickland, Asa Cooper and Petty, Jackson and Pang, Richard Yuanzhe and Dirani, Julien and Michael, Julian and Bowman, Samuel R},
  booktitle={First Conference on Language Modeling},
  year={2024}
}

@article{math,
  title={Measuring mathematical problem solving with the math dataset},
  author={Hendrycks, Dan and Burns, Collin and Kadavath, Saurav and Arora, Akul and Basart, Steven and Tang, Eric and Song, Dawn and Steinhardt, Jacob},
  journal={arXiv preprint arXiv:2103.03874},
  year={2021}
}

@misc{xtuner,
    title={XTuner: A Toolkit for Efficiently Fine-tuning LLM},
    author={XTuner Contributors},
    howpublished = {\url{https://github.com/InternLM/xtuner}},
    year={2023}
}

@article{sglang,
  title={Sglang: Efficient execution of structured language model programs},
  author={Zheng, Lianmin and Yin, Liangsheng and Xie, Zhiqiang and Sun, Chuyue Livia and Huang, Jeff and Yu, Cody Hao and Cao, Shiyi and Kozyrakis, Christos and Stoica, Ion and Gonzalez, Joseph E and others},
  journal={Advances in neural information processing systems},
  volume={37},
  pages={62557--62583},
  year={2024}
}

@article{verl,
  title   = {HybridFlow: A Flexible and Efficient RLHF Framework},
  author  = {Guangming Sheng and Chi Zhang and Zilingfeng Ye and Xibin Wu and Wang Zhang and Ru Zhang and Yanghua Peng and Haibin Lin and Chuan Wu},
  year    = {2024},
  journal = {arXiv preprint arXiv: 2409.19256}
}

@article{timebill,
  title={TimeBill: Time-Budgeted Inference for Large Language Models},
  author={Fan, Qi and Zou, An and Ma, Yehan},
  journal={arXiv preprint arXiv:2512.21859},
  year={2025}
}

@article{latencytts,
  title={Faster and Better LLMs via Latency-Aware Test-Time Scaling},
  author={Wang, Zili and Zhang, Tianyu and Bai, Haoli and Hou, Lu and Yu, Xianzhi and Liu, Wulong and Xiang, Shiming and Zhu, Lei},
  journal={arXiv preprint arXiv:2505.19634},
  year={2025}
}

@article{agenttts,
  title={AgentTTS: Large language model agent for test-time compute-optimal scaling strategy in complex tasks},
  author={Wang, Fali and Liu, Hui and Dai, Zhenwei and Zeng, Jingying and Zhang, Zhiwei and Wu, Zongyu and Luo, Chen and Li, Zhen and Tang, Xianfeng and He, Qi and others},
  journal={arXiv preprint arXiv:2508.00890},
  year={2025}
}


\clearpage
\appendix
\section{Experimental Details}
\label{sec:appendix_exp_details}

\subsection{Details for Timely-Eval}
\label{subsec:timely_eval_details}

In this section, we first introduce the implementation details of \textsc{Timely-Eval}. Specifically, we elaborate on the evaluation protocols for three distinct benchmarks.

\subsubsection{General Benchmarks}
\label{subsubsec:general_benchmarks}

For general benchmarks, we provide the following system prompt to the model to enable time-aware strategic reasoning:

\begin{tcolorbox}[
    colback=gray!5,
    colframe=gray!50,
    title=System Prompt for Time-Aware Strategic Reasoner,
    fonttitle=\bfseries\small,
    fontupper=\small,
    arc=2pt,
    outer arc=2pt
]
Act as a Time-Aware Strategic Reasoner. Your objective is to conquer complex challenges within a strictly enforced time window. 

You must treat Time as your most critical resource. 
1. \textbf{Initial Assessment}: Immediately gauge the problem's complexity against the remaining time. 
2. \textbf{Cognitive Budgeting}: Allocate specific time slots for analyzing, utilizing tools, and synthesizing the final answer. Do not over-invest in low-value details.
3. \textbf{Dynamic Adjustment}: If a reasoning path proves too time-consuming, decisively pivot to a more efficient heuristic or alternative method. 
4. \textbf{Convergence}: You must guarantee a complete conclusion before the deadline. A partial perfect derivation is a failure; a complete, logically sound conclusion is success.

\textbf{Output Requirements}:
- Wrap your condensed final insights in \texttt{<summary>...</summary>} tags.
- Record your actual elapsed duration in \texttt{<conclusion>total duration: \{time\} seconds</conclusion>} tags. Your final answer should be wrapped in \texttt{<answer>} and \texttt{</answer>} tags, for example, \texttt{<answer>$\backslash$boxed\{100\}</answer>}.

\textbf{Tools}:
You may call functions to assist with the user query.
Available tool: \texttt{get\_duration()} - Gets the total elapsed time (in seconds).
\end{tcolorbox}

In our evaluation of the Qwen3 series models, all experiments were consistently conducted on hardware equipped with 8$\times$NVIDIA H200 GPUs. To determine the time constraints for each benchmark, we first performed a baseline inference pass for every case without any time limits. We then established the target time limit by multiplying this baseline duration by a designated coefficient.

\begin{table}[htbp]
\centering
\small
\caption{Full list of 57 games in the Jericho environment.}
\label{tab:jericho_games}
\begin{tabular}{l l l l}
\hline
905 & acorncourt & advent & adventureland \\
afflicted & anchor & awaken & balances \\
ballyhoo & curses & cutthroat & deephome \\
detective & dragon & enchanter & enter \\
gold & hhgg & hollywood & huntdark \\
infidel & inhumane & jewel & karn \\
lgop & library & loose & lostpig \\
ludicorp & lurking & moonlit & murdac \\
night & omniquest & partyfoul & pentari \\
planetfall & plundered & reverb & seastalker \\
sherlock & snacktime & sorcerer & spellbrkr \\
spirit & temple & theatre & trinity \\
tryst205 & weapon & wishbringer & yomomma \\
zenon & zork1 & zork2 & zork3 \\
ztuu & & & \\
\hline
\end{tabular}
\end{table}

\begin{table*}[ht]
\centering
\small
\begin{tcolorbox}[
    colback=white,
    colframe=black!70,
    title=Interaction Example: TimelyLM-8B on MATH Dataset,
    fonttitle=\bfseries,
    arc=0pt
]
\textbf{User}: The volume of the cylinder shown is $45\pi$ cubic cm. What is the height in centimeters of the cylinder? [asy] size(120); draw(shift(2.2,0)*yscale(0.3)*Circle((0,0), 1.2)); ... label("$r=3$",midpoint((2.2,0)--(3.4,0)),N); [/asy] \\
Please answer the question within \textbf{5.31 seconds}. \\
\rule{\linewidth}{0.4pt} \\
\textbf{Assistant}: \texttt{<think>} I'll start by assessing the problem... The volume is $V = \pi r^2 h$. We have $V = 45\pi$ and $r = 3$. \texttt{</think>} \\
\texttt{<tool\_call>} \{"name": "get\_duration", "arguments": \{\}\} \texttt{</tool\_call>} \\
\rule{\linewidth}{0.4pt} \\
\textbf{User}: \texttt{<tool\_response>} 0.54 seconds. \texttt{</tool\_response>} \\
\rule{\linewidth}{0.4pt} \\
\textbf{Assistant}: \texttt{<think>} Good, I have about 4.77 seconds left. $45\pi = \pi \cdot 3^2 \cdot h \Rightarrow 45 = 9h \Rightarrow h = 5$. \texttt{</think>} \\
\texttt{<tool\_call>} \{"name": "get\_duration", "arguments": \{\}\} \texttt{</tool\_call>} \\
\rule{\linewidth}{0.4pt} \\
\textbf{User}: \texttt{<tool\_response>} 1.36 seconds. \texttt{</tool\_response>} \\
\rule{\linewidth}{0.4pt} \\
\textbf{Assistant}: \texttt{<think>} Still plenty of time. The calculation is correct. \texttt{</think>} \\
\texttt{<summary>} The volume is $V = \pi r^2 h$. With $V = 45\pi$ and $r = 3$, we have $45\pi = 9\pi h$, yielding $h = 5$ cm. \texttt{</summary>} \\
\texttt{<conclusion>} total duration: 1.36 seconds \texttt{</conclusion>} \\
\texttt{<answer>}$\boxed{5}$\texttt{</answer>}
\end{tcolorbox}
\captionof{table}{A full reasoning trace of TimelyLM-8B. The model strategically monitors its remaining time using the \texttt{get\_duration} tool to ensure completion within the 5.31s limit.}
\label{tab:math_example}
\end{table*}

\subsubsection{Interactive Games}
\label{subsubsec:interactive_games}

For the interactive games task, our environment is built upon the Jericho framework, which supports 57 interactive fiction games. Models are given the initial observation of the environment, and asked to gain as much score as possible.


The training split specifically includes the following 16 games: \textit{acorncourt, advent, adventureland, anchor, awaken, balances, ballyhoo, curses, cutthroat, deephome, detective, dragon, enchanter, hhgg, hollywood, and zork1}. For these games, the system prompt provided to the model is as follows:

\begin{tcolorbox}[
    colback=gray!5,
    colframe=gray!50,
    title=System Prompt for Interactive Games,
    fonttitle=\bfseries\small,
    fontupper=\small,
    arc=2pt
]
Act as a Time-Aware Strategic Reasoner. Your objective is to conquer complex challenges within a strictly enforced time window. 

You must treat Time as your most critical resource. 
1. \textbf{Initial Assessment}: Gauge complexity vs remaining time. 
2. \textbf{Cognitive Budgeting}: Allocate specific time slots. 
3. \textbf{Dynamic Adjustment}: Pivot to efficient heuristics if needed. 
4. \textbf{Convergence}: Guarantee a conclusion before the deadline. 

\textbf{Output Requirements}:
- Record duration: \texttt{<conclusion>total duration: \{time\} seconds</conclusion>}
- Summarize score: \texttt{<score>\{your\_score\}</score>}
\end{tcolorbox}

The model uses the following tools to interact with the engine:
\begin{itemize}[leftmargin=*, noitemsep]
    \item \texttt{step(action)}: Execute game action.
    \item \texttt{get\_available\_actions()}: Get valid moves.
    \item \texttt{get\_score()}: Check current score.
    \item \texttt{get\_max\_score()}: Check goal score.
    \item \texttt{end\_game()}: Terminate session.
\end{itemize}

Table~\ref{tab:zork_trace} demonstrates the initial strategic reasoning of \textsc{TimelyLM-8B} on \textit{Zork1}.

\subsubsection{Machine Learning Tasks}
\label{subsubsec:ml_tasks}

As the third component of \textsc{Timely-Eval}, we select four representative machine learning (ML) tasks: \textit{leaf\_classification}, \textit{spaceship\_titanic}, \textit{random\_acts\_of\_pizza}, and \textit{detecting\_insults\_in\_social\_commentary}. In these tasks, the model is required to perform end-to-end data science reasoning, including data preprocessing, model selection, training, and evaluation.

The system prompt for ML tasks is designed to emphasize time efficiency and rapid execution:

\begin{tcolorbox}[
    colback=gray!5,
    colframe=gray!50,
    title=System Prompt for Machine Learning Tasks,
    fonttitle=\bfseries\small,
    fontupper=\small,
    arc=2pt
]
Act as a Time-Aware Strategic Reasoner. Your objective is to conquer machine learning reasoning tasks within a strictly enforced time window. 

You must treat Time as your most critical resource. 
1. \textbf{Initial Assessment}: Immediately gauge complexity against remaining time. 
2. \textbf{Cognitive Budgeting}: Allocate time slots for analysis and coding. 
3. \textbf{Dynamic Adjustment}: Pivot to efficient solutions if a method is too slow. 
4. \textbf{Convergence}: Guarantee a complete conclusion before the deadline. 

\textit{Instructions}: Wrap your code in \texttt{<code>...</code>} tags. Avoid long-running code or cross-validation.
\textbf{Output Requirements}:
- \texttt{<conclusion>total duration: \{time\} seconds</conclusion>}
- \texttt{<accuracy>\{accuracy\_float\}</accuracy>}
\end{tcolorbox}

In each round, the model generates an implementation plan and the corresponding Python code. The code is then executed via a specialized tool: \texttt{execute\_code\_and\_get\_duration(code)}. This tool returns the execution \texttt{stdout}, the final evaluation accuracy, and the time consumed.

Table~\ref{tab:ml_example} illustrates a complete execution trace of \textsc{TimelyLM-8B} on the \textit{leaf classification} task.


\subsubsection{Inference and Evaluation Setup}
\label{subsubsec:evaluation_details}

During the evaluation phase, all models were deployed on a hardware cluster equipped with 8$\times$NVIDIA H200 GPUs. For high-throughput inference, we employed the SGLang framework \citep{sglang}.

To balance efficiency and stability, the concurrency level (number of parallel requests) for model calls was set to 16 across all benchmarks. For tasks that exhibit inherent variance—specifically \textbf{Interactive Games} and \textbf{Machine Learning Tasks}—we evaluated each individual case across 8 independent runs. The reported results for these categories represent the aggregate performance across these iterations to ensure statistical significance and mitigate the impact of stochastic generation.

\subsubsection{Time Budget Allocation Strategy}
\label{subsubsec:time_budget_strategy}

\paragraph{General Benchmarks} For general tasks, the time budget is determined on a \textbf{per-case basis}. Let $x$ denote the duration a model requires to complete a specific case during a baseline inference pass (without time constraints). The enforced time budget for that particular case is then defined as $n \times x$, where $n$ is a configurable scaling factor. This approach ensures that the difficulty of each individual problem is intrinsically accounted for.

\paragraph{Interactive Games and ML Tasks} In contrast, for Interactive Games and Machine Learning Tasks, we employ a \textbf{step-based budgeting strategy}. The total time budget is calculated by multiplying a measured average duration per step (derived from baseline performance) by a fixed constant. This ensures a consistent pressure on the model's strategic planning and tool-calling efficiency throughout the multi-turn interaction or the entire coding pipeline.

\begin{table*}[ht]
\centering
\small
\begin{tcolorbox}[
    colback=white,
    colframe=black!70,
    title=Reasoning Trace: TimelyLM-8B on Zork1 (Steps 0--3),
    fonttitle=\bfseries,
    arc=0pt
]
\textbf{System}: [Rules and Tool Definitions] \\
\textbf{User}: Game started. Observation: \textit{West of House. You are standing in an open field west of a white house... There is a small mailbox here.} The time limit is \textbf{47.99 seconds}. \\
\rule{\linewidth}{0.4pt} \\
\textbf{Assistant (Step 0)}: \texttt{<think>} I have 47.99s to maximize my score. Let me get the max score and available actions. \texttt{</think>} \\
\texttt{<tool\_call>} \{"name": "get\_max\_score"\} \texttt{</tool\_call>} \\
\textbf{Tool Response}: The max score is 350. You have played for 3.6 seconds. \\
\rule{\linewidth}{0.4pt} \\
\textbf{Assistant (Step 1)}: \texttt{<tool\_call>} \{"name": "get\_available\_actions"\} \texttt{</tool\_call>} \\
\textbf{Tool Response}: Available actions: ['open mailbox', 'north', 'south', 'west']. You have played for 4.1 seconds. \\
\rule{\linewidth}{0.4pt} \\
\textbf{Assistant (Step 2)}: \texttt{<think>} Max score 350. Time $\approx$ 44s. Let me check my score before starting. \texttt{</think>} \\
\texttt{<tool\_call>} \{"name": "get\_score"\} \texttt{</tool\_call>} \\
\textbf{Tool Response}: Your current score is: 0. You have played for 4.6 seconds. \\
\rule{\linewidth}{0.4pt} \\
\textbf{Assistant (Step 3)}: \texttt{<think>} Time is critical. I'll examine the mailbox first. \texttt{</think>} \\
\texttt{<tool\_call>} \{"name": "step", "arguments": \{"action": "open mailbox"\}\} \texttt{</tool\_call>} \\
\textbf{Tool Response}: Opening the small mailbox reveals a leaflet. Played for 6.0 seconds.
\end{tcolorbox}
\captionof{table}{Initial steps of \textsc{TimelyLM-8B} in Zork1. The model uses the first few seconds to calibrate its goal.}
\label{tab:zork_trace}
\end{table*}

\begin{table*}[b!]
\centering
\small
\begin{tcolorbox}[
    colback=white,
    colframe=gray!80!black,  
    title=ML Reasoning Trace: TimelyLM-8B on Leaf Classification,
    fonttitle=\bfseries,
    colbacktitle=gray!20!black, 
    arc=2pt,
    outer arc=2pt
]
\textbf{User}: The objective is to identify 99 species of plants using shape, margin, and texture features. You must produce \texttt{submission.csv} with predicted probabilities. \\
Data: \texttt{train.csv} (891 rows), \texttt{test.csv} (99 rows). \\
Please finish the task within \textbf{5.32 seconds}. \\
\rule{\linewidth}{0.4pt} \\
\textbf{Assistant}: \texttt{<think>} I'll use a Random Forest classifier. It handles multi-class tabular data effectively and trains quickly. I will: (1) Load data, (2) Encode labels, (3) Train a model with \texttt{n\_jobs = -1}, (4) Generate probabilities. \texttt{</think>}

\begin{lstlisting}[language=Python, basicstyle=\ttfamily\color{blue!60!black}\scriptsize]
import pandas as pd
from sklearn.ensemble import RandomForestClassifier
from sklearn.preprocessing import LabelEncoder

# Load and Prepare
train_df = pd.read_csv('./data/public/train.csv')
test_df = pd.read_csv('./data/public/test.csv')
X = train_df.drop(['id', 'species'], axis=1)
y = LabelEncoder().fit_transform(train_df['species'])

# Train and Predict
clf = RandomForestClassifier(n_estimators=100, random_state=42, n_jobs=-1)
clf.fit(X, y)
X_test = test_df.drop(['id'], axis=1)
probs = clf.predict_proba(X_test)

# Submit
sub = pd.DataFrame(probs, columns=label_encoder.classes_)
sub.insert(0, 'id', test_df['id'])
sub.to_csv('./submission.csv', index=False)
\end{lstlisting}

\texttt{<tool\_call>} \{"name": "execute\_code\_and\_get\_duration", "arguments": \{...\}\} \texttt{<tool\_call>} \\
\rule{\linewidth}{0.4pt} \\
\flushleft
\textbf{Tool Response}: Code execution succeeded. Stdout: Submission file created successfully! \\
\textbf{Evaluation accuracy}: 0.9394. You have spent \textbf{5.22 seconds}. \\
\rule{\linewidth}{0.4pt} \\
\textbf{Assistant}: \texttt{<conclusion>} total duration: 5.22 seconds \texttt{</conclusion>} \\
\texttt{<accuracy>} 0.9394 \texttt{</accuracy>}
\end{tcolorbox}
\caption{An example of \textsc{TimelyLM-8B} solving an ML task. The model selects a computationally efficient Random Forest model to ensure the entire pipeline (loading, training, and prediction) completes within the strict 5.32s limit.}
\label{tab:ml_example}
\end{table*}


\subsection{Details for Timely-RL}
\label{subsec:timely_rl_details}

In this section, we provide a comprehensive description of the implementation and training details for \textsc{Timely-RL}. Our RL framework is built upon the \texttt{VeRL} \citep{verl} and \texttt{InternBootCamp} frameworks. We designed three categories of tools and corresponding RL pipelines tailored to the three task types. For the sake of clarity, we restate the unified reward function introduced in Section~\ref{sec:timelyrl} here:

\begin{equation}
\label{eq:appendix_reward_formula}
R(t, r) = 
\begin{cases}
0, & t > T_{\max} \\
r_{\text{f}}, & t \le T_{\max}, r = 0 \\
r_{\text{f}} + r + \lambda U(t), & t \le T_{\max}, r > 0
\end{cases}
\end{equation}
where $r_{\text{f}}$ is a fixed reward for correct formatting and timely completion, and $U(t)$ represents the model's utilization rate of time cost, calculated as:
\begin{equation}
U(t) = \sin\left( \frac{\pi}{2} \cdot \min\left(\frac{t}{T_{\text{max}}}, 1\right) \right)
\end{equation}

\subsubsection{Timer Tool}
\label{subsubsec:timer_tool}

We implemented a \textit{Timer Tool} that supports a single function call: \texttt{get\_duration()}. To simulate diverse execution environments, the tool's initialization allows for specific time scaling coefficients and stochastic perturbation factors (jitter). When the model calls this tool, the returned duration is the product of the actual elapsed time and a designated coefficient, plus a random noise component. This design enables \textsc{Timely-RL} to generalize across different computational resources and varying inference speeds.

\subsubsection{General Reasoning RL}
\label{subsubsec:general_reasoning_rl}

For general reasoning tasks, the \textit{Timer Tool} serves as the primary interface for time-awareness. We deployed a dedicated service to manage timers for different cases. Before each \texttt{VeRL} rollout session, every case is registered with this service to ensure that the timing for each instance is strictly independent and isolated. 

During the RL process, the model is encouraged to call \texttt{get\_duration()} to monitor its own reasoning progress. Following the reward formulation in Eq.~\ref{eq:appendix_reward_formula}, for General Reasoning tasks, we set the parameters as follows: the task reward $r = 0.5$ (for correct answers), the formatting reward $r_{\text{f}} = 0.1$, and the time-utilization weight $\lambda = 0.4$.

\subsubsection{Interactive Games RL}
\label{subsubsec:interactive_games_rl}

For the interactive games task, we deploy a specialized service to host the Jericho game engine. Specifically, an isolated game instance is initialized prior to each rollout session. The model interacts with this environment by invoking designated tools until the game reaches a terminal state or the model explicitly calls the \texttt{end\_game} tool. We integrate the \textit{Timer Tool} (described in Section~\ref{subsubsec:timer_tool}) directly into this service, ensuring that the model receives both environmental feedback and the total elapsed time in a single turn.

Critically, we differentiate between \textbf{inference time} (the latency of model generation) and \textbf{tool call time} (the latency of environment execution, including simulated tool delays). To simulate diverse reasoning speeds, we apply a scaling coefficient to the actual inference time, while the tool call duration remains unadjusted. This separation allows us to better analyze the model's strategic behavior under varied computational constraints. For interactive games, r is calculated as follows: 

\begin{equation}
    r = (\frac{s_{curr}}{s_{max}})^{1/3}
\end{equation}
where $s_{curr}$ is the score gained in this game, and $s_{max}$ denotes the maximum score of the game.

\subsubsection{Machine Learning Tasks RL}
\label{subsubsec:ml_tasks_rl}

Lastly, for machine learning tasks, we deploy a dedicated code execution service to process the model-generated scripts. Unlike interactive games, ML tasks do not require maintaining complex environmental state transitions; therefore, we simply instantiate a separate \textit{Timer Tool} for each case to monitor execution duration. 

In this category, the performance component $r$ in Eq.~\ref{eq:appendix_reward_formula} corresponds to the evaluation metric (accuracy) achieved by the model's code on the test set. Since all four selected ML benchmarks are classification tasks, $r$ is naturally a floating-point value within the range $[0, 1]$. To balance the total reward magnitude, we define the performance contribution to the final reward as $0.5 \times r$.



\subsection{Prompts for ML Tasks}
\label{subsec:prompts_ml}

We designed individual task prompts for the four selected machine learning benchmarks. The detailed prompts for each task are provided below.

\small 

\begin{tcolorbox}[
    colback=gray!5, 
    colframe=gray!50, 
    title=Task 1: Detecting Insults in Social Commentary, 
    fonttitle=\bfseries,
    fontupper=\raggedright 
]
Please first outline your proposed solution in natural language, followed by a single markdown code block. 

The task description is: The challenge is to detect when a comment from a conversation would be considered insulting to another participant. The goal is to create a generalizable classifier for real-time moderation.

\textbf{Note}:
\begin{itemize}[leftmargin=1em, nosep]
    \item Datasets \texttt{train.csv} and \texttt{test.csv} are in \texttt{./data/public/}.
    \item You MUST produce \texttt{./submission.csv} on the test split.
    \item Predict probability (0 to 1). Evaluated using AUC.
    \item \texttt{./submission.csv} format: \\
    \texttt{Insult,Date,Comment}
\end{itemize}
\end{tcolorbox}

\begin{tcolorbox}[
    colback=gray!5, 
    colframe=gray!50, 
    title=Task 2: Leaf Classification, 
    fonttitle=\bfseries,
    fontupper=\raggedright 
]
Please first outline your proposed solution in natural language, followed by a single markdown code block. 

The task description is: Use binary leaf images and extracted features (shape, margin, texture) to identify 99 species of plants. 

\textbf{Note}:
\begin{itemize}[leftmargin=1em, nosep]
    \item Datasets: \texttt{train.csv}, \texttt{test.csv}, and \texttt{images/} in \texttt{./data/public/}.
    \item Task: Predict probabilities for 99 species. Evaluated using multi-class logarithmic loss.
    \item Predicted probabilities should be clipped: \\ 
    $\max(\min(p, 1-10^{-15}), 10^{-15})$.
    \item \texttt{./submission.csv} format: \\
    \texttt{id, Acer\_Capillipes, Acer\_Circinatum, ...}
\end{itemize}
\end{tcolorbox}

\begin{tcolorbox}[
    colback=gray!5, 
    colframe=gray!50, 
    title=Task 3: Random Acts of Pizza (RAOP), 
    fonttitle=\bfseries,
    fontupper=\raggedright 
]
First outline your proposed solution in natural language, followed by a single markdown code block. 

The task description is: Predict which textual requests for pizza from the Reddit community will be successful based on request text and meta-data.

\textbf{Note}:
\begin{itemize}[leftmargin=1em, nosep]
    \item Datasets: \texttt{train.json} and \texttt{test.json} in \texttt{./data/public/}.
    \item Task: Predict probability of receiving pizza. Evaluated on ROC AUC.
    \item \texttt{./submission.csv} format: \\
    \texttt{request\_id, requester\_received\_pizza}
\end{itemize}
\end{tcolorbox}

\begin{tcolorbox}[
    colback=gray!5, 
    colframe=gray!50, 
    title=Task 4: Spaceship Titanic, 
    fonttitle=\bfseries,
    fontupper=\raggedright 
]
First outline your proposed solution in natural language, followed by a code block.

The task description is: Predict which passengers were transported to an alternate dimension after the \textit{Spaceship Titanic} collided with a spacetime anomaly.

\textbf{Note}:
\begin{itemize}[leftmargin=1em, nosep]
    \item Datasets: \texttt{train.csv} and \texttt{test.csv} in \texttt{./data/public/}.
    \item Task: Achieve best classification accuracy on test split.
    \item \texttt{./submission.csv} format: \\
    \texttt{PassengerId, Transported}
\end{itemize}
\end{tcolorbox}
\normalsize

\subsection{Prompts for Interactive Games}
\label{subsec:prompts_games}

In this section, we present the initial observations and prompts received by the model during the first step of several representative games in the Jericho environment. Each task provides the model with a \texttt{max\_score} target and a textual description of the starting environment.

\small 

\begin{tcolorbox}[
    colback=black!5, 
    colframe=gray!50, 
    title=Game: Zork1 (Target Score: 350), 
    fonttitle=\bfseries,
    fontupper=\fontfamily{cmtt}\selectfont\raggedright\footnotesize
]
Game started. Initial observation:\\
Copyright (c) 1981, 1982, 1983 Infocom, Inc. All rights reserved.\\
ZORK is a registered trademark of Infocom, Inc.\\
Revision 88 / Serial number 840726\\
\vspace{0.5em}
West of House\\
You are standing in an open field west of a white house, with a boarded front door. There is a small mailbox here.
\end{tcolorbox}

\begin{tcolorbox}[
    colback=black!5, 
    colframe=gray!50, 
    title=Game: Detective (Target Score: 360), 
    fonttitle=\bfseries,
    fontupper=\fontfamily{cmtt}\selectfont\raggedright\footnotesize
]
Game started. Initial observation:\\
Type "help" for more information about this version\\
Detective. By Matt Barringer. Ported by Stuart Moore.\\
Release 1 / Serial number 000715 / Inform v6.21\\
\vspace{0.5em}
<< Chief's office >>\\
You are standing in the Chief's office. He is telling you "The Mayor was murdered yesterday night at 12:03 am. I want you to solve it before we get any bad publicity or the FBI has to come in." "Yessir!" You reply. He hands you a sheet of paper. Once you have read it, go north or west.\\
\vspace{0.5em}
You can see a piece of white paper here.\\
Your score has just gone up by ten points.
\end{tcolorbox}

\begin{tcolorbox}[
    colback=black!5, 
    colframe=gray!50, 
    title=Game: Advent (Target Score: 350), 
    fonttitle=\bfseries,
    fontupper=\fontfamily{cmtt}\selectfont\raggedright\footnotesize
]
Game started. Initial observation:\\
Welcome to Adventure! (Please type HELP for instructions.)\\
ADVENTURE. The Interactive Original.\\
By Will Crowther (1976) and Don Woods (1977).\\
\vspace{0.5em}
At End Of Road\\
You are standing at the end of a road before a small brick building. Around you is a forest. A small stream flows out of the building and down a gully.
\end{tcolorbox}

\begin{tcolorbox}[
    colback=black!5, 
    colframe=gray!50, 
    title=Game: Enchanter (Target Score: 400), 
    fonttitle=\bfseries,
    fontupper=\fontfamily{cmtt}\selectfont\raggedright\footnotesize
]
Game started. Initial observation:\\
"Krill's evil must be unmade," he begins... "Only then may his vast evil be lessened or, with good fortune, destroyed."\\
\vspace{0.5em}
ENCHANTER. Infocom interactive fiction.\\
Copyright (C) 1983, 1984 by Infocom, Inc. All rights reserved.\\
\vspace{0.5em}
Fork\\
You stand at a point of decision on a road which makes a wide fork to the northeast and southeast, circling the base of the Lonely Mountain, which looms high overhead to the east. A very long and winding road starts here and stretches out of sight to the west through low, smoky hills. The sun is rising over the lands to the east.
\end{tcolorbox}

\normalsize

\section{Distillation Details}
\label{sec:appendix_distillation}

In this section, we provide the implementation details for the distillation process used to generate cold-start data, specifically focusing on the General Reasoning tasks.

To provide a high-quality cold-start foundation for \textsc{Timely-RL}, we developed a distillation pipeline using queries sampled from the AM-1.4M dataset. We utilized \texttt{Qwen3-235B-Instruct-2507} as the teacher model, equipped with a \textit{Timer Tool} and instructed to complete tasks accurately within strictly enforced deadlines. To ensure the model generalizes across various hardware scenarios and reasoning speeds, we randomly sampled different time limits and timer acceleration coefficients during data synthesis.

For response verification, we employed a two-stage checking process:
\begin{enumerate}[leftmargin=*, noitemsep]
    \item \textbf{Rule-based Verification}: We first used the \texttt{math-verify}\footnote{\url{https://github.com/huggingface/Math-Verify}} library to perform symbolic and rule-based correctness checks.
    \item \textbf{Model-based Verification}: Cases that failed rule-based checks were further evaluated by \texttt{Qwen3-235B}. Each case was checked 3 times; if all three evaluations passed, the solution was deemed accurate.
\end{enumerate}

The system prompt provided to the teacher model during data generation is shown below:

\begin{tcolorbox}[
    colback=gray!5, 
    colframe=gray!50, 
    title=System Prompt for Distillation (Teacher Model), 
    fonttitle=\bfseries\small,
    fontupper=\small,
    arc=2pt
]
Act as a Time-Aware Strategic Reasoner. Your objective is to conquer complex challenges within a strictly enforced time window. 

You must treat Time as your most critical resource. 
1. \textbf{Initial Assessment}: Immediately gauge complexity against remaining time. 
2. \textbf{Cognitive Budgeting}: Allocate time slots for analysis and tools. 
3. \textbf{Dynamic Adjustment}: Pivot to efficient heuristics if needed. 
4. \textbf{Convergence}: Guarantee a complete conclusion before the deadline. 

\textbf{Output Requirements}:
- Wrap insights in \texttt{<summary>...</summary>} tags.
- Record duration in \texttt{<conclusion>total duration: \{time\} seconds</conclusion>}. 
- Wrap final answer in \texttt{<answer>$\backslash$boxed\{...\}</answer>}.
- \textit{Note}: Keep final conclusions brief. Split reasoning into multiple short steps and call the \texttt{get\_duration} tool frequently to avoid estimation errors.
\end{tcolorbox}

For the verification stage, the system prompt for the \textbf{judger model} is as follows:

\begin{tcolorbox}[
    colback=gray!5, 
    colframe=gray!50, 
    title=System Prompt for Judger Model, 
    fonttitle=\bfseries\small,
    fontupper=\small,
    arc=2pt
]
You are an expert agent evaluating the timely reasoning tasks. You will be given the final answer and the correct answer, and you should evaluate the final answer and output the evaluation result.

Please answer only \textbf{yes} or \textbf{no} in the following format:
\texttt{<answer>yes or no</answer>}
\end{tcolorbox}

\section{Broader Impacts}
\label{sec:appendix_impacts}

In this section, we discuss the broader implications, ethical considerations, and reproducibility of our work.

\paragraph{Ethics Statement} 
All datasets utilized in this study are obtained from publicly available sources and are free for academic research. Our data synthesis and evaluation processes do not involve the collection or processing of any sensitive personally identifiable information (PII). Consequently, we do not foresee any significant privacy risks associated with the data used in this work.

\paragraph{Reproducibility} 
We are committed to open-source and reproducible research. To facilitate the reproduction of our experimental results, we will release our complete source code, the synthesized distillation datasets, and the \textsc{Timely-Eval} evaluation framework upon publication. Detailed implementation parameters and hardware configurations are also documented in the preceding sections of this appendix.

\paragraph{Contributions and Perspective} 
The primary contribution of this research is the introduction of a novel perspective on \textit{test-time scaling} within agentic scenarios. We posit that test-time in agentic contexts should be directly coupled with physical wall-clock time rather than purely symbolic computation steps. By providing a theoretical framework and designing the \textsc{Timely-RL} algorithm, we demonstrate that time-aware strategic reasoning is a distinct capability that can be significantly enhanced through specialized training. We hope this work inspires further research into making autonomous agents more efficient and reliable in time-sensitive real-world environments.




\end{document}